\documentclass[sn-mathphys-num]{sn-jnl}

\usepackage{graphicx}%
\usepackage{multirow}%
\usepackage{amsmath,amssymb,amsfonts}%
\usepackage{amsthm}%
\usepackage{mathrsfs}%
\usepackage[title]{appendix}%
\usepackage{xcolor}%
\usepackage{textcomp}%
\usepackage{manyfoot}%
\usepackage{booktabs}%
\usepackage{algorithm}%
\usepackage{algorithmicx}%
\usepackage{algpseudocode}%
\usepackage{listings}%
\usepackage{wrapfig}

\definecolor{codegreen}{rgb}{0,0.6,0}
\DeclareMathOperator*{\argmin}{arg\,min}

\theoremstyle{thmstyleone}%
\theoremstyle{thmstyletwo}%

\theoremstyle{thmstylethree}%

\begin{document}

\title{Model Debiasing by Learnable Data Augmentation}

\author*[1]{\fnm{Pietro} \sur{Morerio}}\email{pietro.morerio@iit.it}
\equalcont{These authors contributed equally to this work.}
\author[1]{\fnm{Ruggero} \sur{Ragonesi}}\email{ruggero.ragonesi@iit.it}
\equalcont{These authors contributed equally to this work.}

\author[1,2]{\fnm{Vittorio} \sur{Murino}}\email{vittorio.murino@iit.it}

\affil*[1]{\orgdiv{PAVIS}, \orgname{Istituto Italiano di Tecnologia}, \orgaddress{\city{Genova}, \country{Italy}}}

\affil[2]{\orgdiv{Department of Computer Science}, 
\orgname{University of Verona}, \orgaddress{
\city{Verona}, 
\country{Italy}}}

\abstract{Deep Neural Networks are well known for efficiently fitting training data, yet experiencing poor generalization capabilities whenever some kind of bias dominates over the actual task labels, resulting in models learning ``shortcuts''. 
In essence, such models are often prone to learn spurious correlations between data and labels. 
In this work, we tackle the problem of learning from biased data in the very realistic \textit{unsupervised} scenario, i.e., when the bias is unknown. This is a much harder task as compared to the supervised case, where auxiliary, bias-related annotations, can be exploited in the learning process.
This paper proposes a novel 2-stage learning pipeline featuring a data augmentation strategy able to regularize the training.  
First, biased/unbiased samples are identified by training over-biased models. 
Second, such subdivision (typically noisy) is exploited within a data augmentation framework, properly combining the original samples while \textit{learning} mixing parameters, which has a regularization effect. 
Experiments on synthetic and realistic biased datasets show state-of-the-art classification accuracy, outperforming competing methods, ultimately proving robust performance on both biased and unbiased examples. Notably, being our training method totally agnostic to the level of bias, it also positively affects performance for any, even apparently unbiased, dataset, thus improving the model generalization regardless of the level of bias (or its absence) in the data.
}


\maketitle

\section{Introduction}

The learning capabilities of deep neural networks are nowadays universally recognized, especially in classification tasks, where deep architectures are able to fit large amounts of data, reaching unprecedented performance. Yet, being the learning process purely data-driven, such models can also extremely prone to fail (often with high confidence) whenever test samples are drawn from a different distribution (or domain).

One of the reasons for poor generalization also lies in the possible bias(es) present in the training samples, i.e., whenever significant portions of data show spurious correlations with class labels. This can lead the trained model to learn the so-called ``shortcuts'' to classify data, thus failing to generalize properly \cite{beery_terra_incognita,shortcut_learning}. 
For example, a duck can be classified as such by a biased model due to the presence of blue water in the surroundings, and not for the actual bird shape and appearance. Hence, the very same model will likely fail in case the input image depicts a duck located on land or, vice versa, another bird located on water may be mis-classified as a duck. This kind of ``shortcuts'' are usually learnt since a \textit{vast majority} of the samples are characterized by a spurious bias (e.g., ducks in water), while only a few are unbiased (e.g., ducks on land).

\begin{figure*}[!t]

    \includegraphics[width=1.\textwidth]{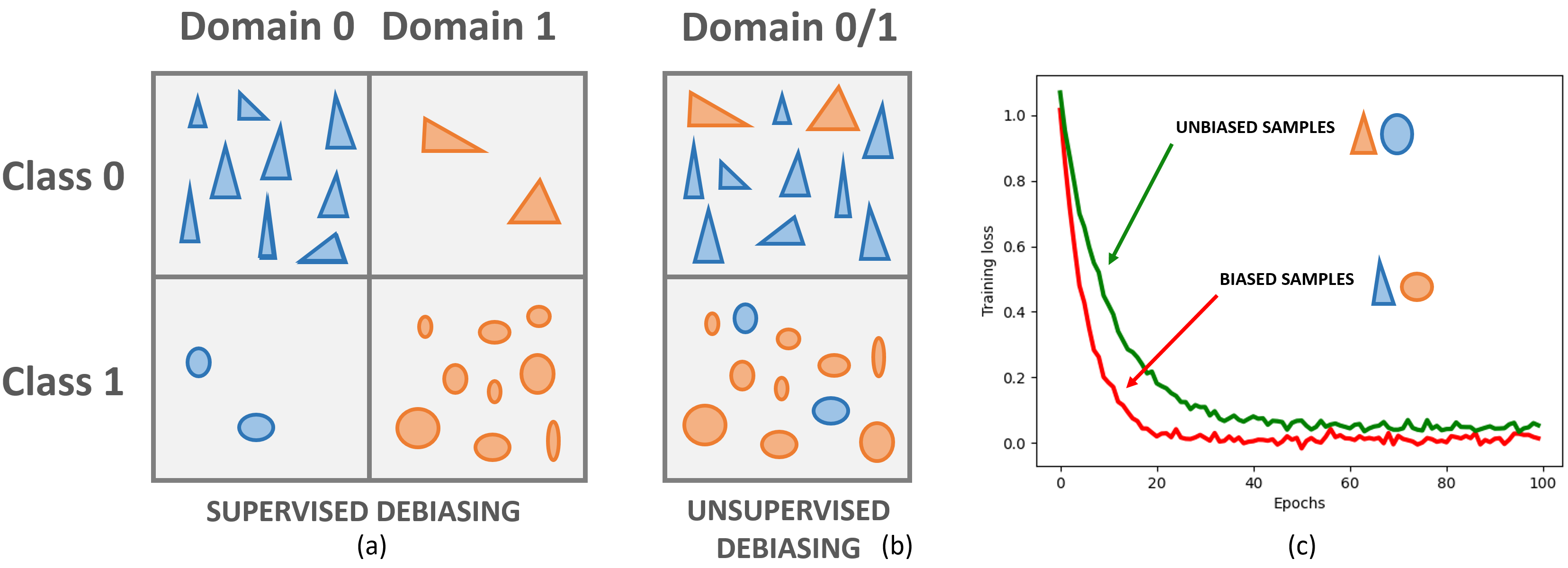}
    \caption{ \footnotesize \textbf{Problem description.} (a) Biased datasets exhibit a class/domain imbalance, namely, one or more classes are mostly observed under one domain, leaving other options under-represented. In the case of supervised debiasing, additional information/annotations regarding the 
    domain distribution are available. 
    (b) In unsupervised debiasing settings, only class labels are available. Possibly, pseudo-labelling can be adopted to fall back into the supervised settings scenario.
    (c) Biased samples are fitted more easily than unbiased ones, as indicated by the different rates at which the average loss decreases. (Best viewed in color). }
    \label{fig:problem_description}
\end{figure*}

There are several approaches to cope with this problem, depending on whether the bias is known or not.
In the \textit{supervised debiasing} settings, a model is optimized in presence of biased data assuming the ground-truth knowledge of the bias.
Such additional auxiliary annotations can be used to drive model optimization towards a data representation invariant to the domain (or attribute), as in \cite{alvi2018turning,kim2019learning,li2019repair,wang2019learning,ragonesi2020learning,group_DRO,clark2019dont,IRM} (see Figure \ref{fig:problem_description}(a) for a visual explanation).  
However, the availability of such additional domain labels is mostly unrealistic in many practical scenarios, as it would require a great effort during data annotation, and in some cases it could even be impossible to obtain since biases could not be directly interpretable, nor perceivable by humans. Hence, the need for methods that can achieve robust generalization in presence of bias, with no additional supervision nor labelling is strongly desirable.

The \textit{unsupervised debiasing} scenario is in fact more realistic and challenging, and is the problem we are addressing here (Figure \ref{fig:problem_description}(b)). Assuming that the ground-truth annotations for the bias are not available, the idea is to attempt to (implicitly) infer such information and exploit it to debias our model. We do so by designing a data augmentation strategy, while properly weighting all the data samples proportionally to their bias level. 
Ultimately, the proposed approach aims at regularizing the training process by counterbalancing the biased samples with more ``neutral'', augmented examples, hence mitigating the learning of the bias, while achieving a successful generalization on the test set, often containing balanced (biased and unbiased) data.

In this paper, we devised a two-stage pipeline to tackle the unsupervised debiasing problem. First, we devise a methodology for binary separation of biased/unbiased samples via a pseudo-labeling approach. Second, equipped with (noisy) bias/unbias pseudo-labels, we approach the problem of learning unbiased representations via a data augmentation strategy: we design a new objective function properly weighting the contribution of the two noisy subsets while learning an augmented version of the samples. Specifically, we generate (augment) new data by linearly interpolating biased and unbiased pseudo-labeled samples thus producing data which are more ``neutral'' than the original biased or unbiased data. This reduces the contribution of spurious correlations, which cause overfitting of biased samples as shown in Fig. \ref{fig:problem_description}(c). 
More specifically, we augment data by adopting a variant of Mixup \cite{mixup}, i.e., a learnable version of it, where the parameters governing the data mixing are learnt, to increase the debiasing effect and boost the method's performance.
It is important to note that, while existing methods \cite{nam2020learning,clark2019dont} are primarily devoted to increasing accuracy on unbiased samples, overlooking the need for keeping high accuracy on biased data as well, we aim instead at achieving high accuracy over both types of data. Notably, the method is also completely agnostic of the presence of bias, as it shows positive effects even when no bias is (apparently) present in the data.

We test our method on benchmarks that both contain synthetically generated and controlled bias (Corrupted CIFAR-10), and more realistic (Waterbirds, CelebA and BAR), showing superior performance with respect to state-of-the-art methods.

\noindent In summary, the main contributions of our work are:
\begin{itemize}
\item We deal with the challenging \textit{unsupervised debiasing} problem by proposing a novel two-stage approach to tackle it: fort we propose an initial data sorting technique aimed at segregating biased and unbiased samples; after that, we employ data augmentation to properly regularize the training, limiting the detrimental effect of the bias.

\item Our proposed approach learns unbiased models from biased and unbiased samples, by weighting their contributions through mixing. Specifically, this can be seen as a data augmentation process, consisting in a principled approach to learn the parameters controlling the mixing of the data samples.
By considering augmented (mixed) data in the loss function, we inject a strong regularization signal in the training process, thus compensating the imbalance problem and leading to a more general and unbiased representation learning. 
\item We present an extensive validation of the proposed pipeline on datasets with controlled bias as well as realistic benchmarks, outperforming state-of-the-art approaches by a significant margin.
\end{itemize}

\noindent The rest of the paper is structured as follows. Section \ref{sec:relwork} discusses the related literature and highlights the original contributions of our proposed approach. 
Section \ref{sec:formulation} introduces the problem formulation and the related notation. The initial biased/unbiased data subdivision process is reported in Section \ref{sec:BI}. Section \ref{sec:repr_learn} instead details the 2nd stage consisting in the design of a loss function achieving invariant representation learning based on sample weighting and data augmentation via the learning of the mixing parameters to control the regularization. Sections \ref{sec:experiments} and \ref{sec:ablation} present the results and a thorough ablation analysis, respectively. Finally, Section \ref{sec:conclusion} draws conclusions and sketches future research directions.

\section{Related Works}
\label{sec:relwork}

The problem of learning from biased data has been explored in past years mostly in the \textit{supervised} setting, i.e. when labels for the bias factor are available.
Several methods tackled it seeking an invariant data representation to such a known factor. These approaches relied on adversarial learning \cite{zhang2018mitigating,alvi2018turning,kim2019learning}, variational inference \cite{variational_fair_AE,moyer2018NIPS,creager2019flexibly}, Information Theory \cite{ragonesi2020learning}, re-sampling strategies \cite{li2019repair}, or robust optimization \cite{group_DRO}. 
In this context, it is notable to quote Invariant Risk Minimization \cite{IRM}, which seeks an optimal representation  invariant across domains, and EnD \cite{tartaglione_2021}, which proposes to insert an information bottleneck module to  disentangle useful information from the bias.

A more interesting scenario is when bias(es) in the data is unknown, i.e., the \textit{unsupervised} case, which is the case we are addressing \cite{bahng2020learning,CVaR_DRO,nam2020learning,JustTrainTwice}. 

Several approaches have been lately proposed, the most straighforward consisting in first identify the bias and then apply supervised techniques to reduce its impact (e.g., \cite{sagawa2019distributionally, ross2017right, cadene2019rubi, group_DRO, selvaraju2019taking}. 
Unfortunately, these approaches are not always optimal since the bias is typically not easily identifiable, nor sometimes can be clearly associated to a specific attribute of the data. Hence, they likely fails since the debiasing is based on only an approximated, hence inaccurate, estimation of the bias information.

Nevertheless, a large part of these techniques is still based on the initial estimation or the learning of the unknown bias, which is subsequently exploited for the actual debiasing stage in a more robust manner. 
For example, some methods leverage additional supporting models (e.g., \cite{zhang2018mitigating, Teney22, nam2020learning, Kim2022, JustTrainTwice, kim2019learning}), typically in the form of adversarial methods or ensembles of classifiers, to learn the biases in the data, and use them to condition the primary model in order to mitigate their effects.
Specifically, this kind of approaches (e.g., \cite{kim2019learning}) tends to decouple the bias contribution in the feature embeddings, e.g., by minimizing the mutual information between such embeddings and the bias. Leveraging an additional network to predict the bias distribution, a primary network is adversarially trained against the feature embedding network. At convergence, the bias prediction network is not able to predict the bias, while the feature embedding network has successfully ``unlearnt'' it.

Other works (e.g., \cite{Hwang22, nam2020learning, Seo22, Ragonesi_2023_CVPR}) attempt to identify the so-called bias-aligned and bias-conflicting samples (another way to name biased and unbiased samples), followed by an adaptive re-weighting and re-sample procedure regularizing the training. 
The goal here is to debias the training process by properly weighting the individual samples on the basis of the (estimated) level of bias they are affected, so that bias-conflicting examples, usually very few, are generally given more importance with respect to the more numerous bias-aligned data.

Self-supervised or clustering mechanisms \cite{Hong21, Jung21} are also used to promote the aggregation of samples with the same target class, but different bias.  
Generative bias-transformation networks are also used as  translation models to transform the bias in the samples, so that bias-invariant representations can be learnt via contrastive learning. 
Similar ``data transformation'' approaches are also designed, aimed at leveraging the generation of new samples, often under the form of data augmentation, in order to counterbalance the effect of the majority of the biased examples during training. \cite{biaswap, Hwang22}.

Another class of methods tackles the debiasing task considering robust risk minimization \cite{bahng2020learning,CVaR_DRO} in training, i.e., by learning a (debiased) network that is statistically independent from a model trained to be strongly biased by design. Bias amplification methods, such as \cite{Lee23Ampl}, push to an extreme the training of a model to be strongly biased, that is, overfitted to the bias, to split between bias-aligned and conflicted examples. A different approach is proposed in \cite{Shrestha22-1}, which operates at the architectural level, i.e., the network architecture is modified to impose inductive biases that make the network robust to dataset bias. The first inductive bias is imposed in order to use the most little network depth as needed for an individual example. The second inductive bias is instead pointed to use fewer image locations for prediction.

To recap, it can be noted that the debiasing approaches in unsupervised scenarios share similar logical structures while differentiating from the specific methods to identify the bias attributes or estimating bias-aligned or -conflicting samples. Such methods span from the adoption of ensembling or ad-hoc auxiliary models, individual data re-weighting and re-sampling, clustering, the use of contrastive approaches or the design of special networks forcing feature orthogonality, to the generation of new unbiased data (augmentation) aimed at regularizing the training.

Differently from several former methods \cite{Teney22,Kim2022,nam2020learning,bahng2020learning}, our work does not rely on an ensemble of networks to have a reference biased model, neither we perform data upsampling as in \cite{JustTrainTwice}, \cite{nam2020learning} and \cite{li2019repair}, 
nor we consider the confidence of the predictions (as in \cite{biaswap}).
Instead, we pursue an original pseudo-labeling approach to split the dataset in two subsets, biased and unbiased, which we subsequently use together with newly generated (augmented) data to regularize the overall training optimization process and debias the model.
More specifically, we adopt a data augmentation approach combining biased and unbiased samples: inspired by \textit{mixup} \cite{mixup}, we mix samples presenting peculiar features of the bias regime 
(likely representing a shortcut to infer the class) with samples that are deemed not being affected by it, so to produce new, ``neutral'' examples able to counterbalance the bias influence.
This is done by devising an adaptive adversarial mechanism that aims at maximizing the classification performance, while learning the parameters governing the mix of the samples. 
The adversarial mechanism is in fact introduced to challenge the classifier with augmented, difficult to correctly classify, samples, which are generated specifically to reduce the detrimental contribution of the bias during training. By optimizing such data augmentation mechanism by learning the best parameters of the mixing, we generate synthetic data  expected to better break the spurious correlations that affect the original data, eventually leading to a stronger  regularization and effectively debias the model during the training process. 

\section{Problem formulation}
\label{sec:formulation}

\noindent We consider supervised classification problems with a training set 
$\mathcal{D}_{train}=\{x_k,y_k,d_k\}_{k=1}^N$, 
where $x_k$ are raw input data (e.g., images), $y_k$ class labels and $d_k$ domain labels. In the case of a biased dataset, $\mathcal{D}_{train}$ has several classes $y^i, i=1,...,C$, which are considered to be observed under different domains $d^j, j=1,...,D$.
In general, $D$ can be different from $C$ but here, for clarity and without losing generality, we consider the case of $D=C$. 
When the majority of samples of any class $y^i$ is observed under a single domain $d^j$, we say that there is spurious correlation between $(y^i,d^j)$, i.e., the dataset has a bias.

We define $\mathcal{D}_{bias}$ as the subset of training samples that exhibit spurious correlations and $\mathcal{D}_{unbias}$ as the subset of samples with under represented pairs. Such subsets are highly imbalanced, i.e. $| \mathcal{D}_{bias} | >>  | \mathcal{D}_{unbias} | $, otherwise, when the bias would not account for the majority of class samples, its impact can be weaker and it can not be considered a bias anymore.
For instance, in a cats vs. dogs classification problem, most of the cats may be observed in an indoor home environment, while most of the dogs may be observed in outdoor scenes. For both classes, very few images are outside their main distribution, and models could identify the domain (indoor/outdoor) as the class.

When domain labels $d$ are not available, nor we have access to other bias information, we are facing the \textit{unsupervised debiasing} problem. Hence, we consider a training set only containing input data and class label, $\mathcal{D}=\{\mathbf{x,y}\}=\{x_k,y_k\}_{k=1}^N$. We consider the task of training a neural network $f_{\theta}$ on $\mathcal{D}$, with parameters $\theta$, which are usually found via Empirical Risk Minimization (ERM), i.e., by minimizing the expected Cross-Entropy loss over the training data:
\begin{equation}
\theta^{*} =  \underset{\theta}{\argmin}
   \frac{1}{N} \sum_{k=1}^N\mathcal{L}_{CE} (y_k, f_{\theta}(x_k)) \\
\label{eq:ERM}
\end{equation}

\noindent In such scenario, when trained via ERM, a model focuses mostly on the more numerous biased samples, underfitting the unbiased ones: this results in a biased model that uses spurious correlations as shortcuts to make inference, instead of correctly learning the class semantics.
In general, $\mathcal{D}_{test}$ follows a data distribution different from $\mathcal{D}_{train}$, i.e. the biased samples may not be the majority, hence, it is important to learn a model that can be efficiently deployed for both biased and unbiased samples.
As also briefly mentioned above, please note that when a class is not observed predominantly under a single domain, the dataset could not be considered to be biased: in fact an ERM model can learn a category's features from a a multitude of possible domains, hence mitigating the learning of spurious correlations.

In general, the notion of \textit{domain} might be very 
fuzzy, being possibly not interpretable or even unknown: domains could in fact lack a manifest semantic meaning. Moreover, there could also be the case of multiple predominant domains for a single class, the opposite case of the same predominant domain shared among a subset of classes, or any scenario in-between. 

Please, also note that, typically, the more domains are involved, the less biased is the scenario, since heterogeneous training data may be sufficient to train a model with standard ERM without resorting to any special algorithm. 
To sum up, we are trying to face a realistic situation in which data samples are acquired without much control other than its class, resulting in a dataset potentially biased in an unknown manner or even unbiased in fortunate cases. 
\\

Our method tackles the unsupervised debiasing problem with a two-stage approach. In the first stage (Section \ref{sec:BI}), we try to separate biased from unbiased samples through a pseudo-labeling algorithm. Equipped with such (noisy) pseudo-labels, we train a model to produce a data representation that can accommodate both biased and unbiased samples regardless of the severity of the data bias (Section \ref{sec:repr_learn}).

In the next section, in order to perform preliminary studies on a dataset with controlled bias, we use as a toy benchmark a modified version of the CIFAR-10 dataset, named Corrupted CIFAR-10, which is a modification of the original dataset \cite{cifar10} that has been introduced in \cite{corrupted_cifar}, containing 50,000 training RGB images and 10 classes. The bias here stems from the fact that each image is corrupted with a specific noise (e.g., Gaussian blur, salt and pepper noise, etc.). Specifically, each class has a privileged type of noise under which it is observed during training (e.g., most of car images are corrupted with motion blur).

\section{Bias Identification} \label{sec:BI}

\noindent In this first stage, our goal is to split the training set $\mathcal{D}$ into two disjoint subsets, $\hat{\mathcal{D}}_{bias}$ and $\hat{\mathcal{D}}_{unbias}$, which should reflect the actual ground-truth $\mathcal{D}_{bias}$ and $\mathcal{D}_{unbias}$. 

In \cite{nam2020learning, biaswap}, it is shown how the biased samples are learnt faster than the unbiased ones: the imbalanced nature of the dataset makes the model more prone to learn first the numerous biased samples and later the unbiased ones.

\begin{figure}[t!]
    
    \centering

    \includegraphics[width=0.66\textwidth,height=0.38\textwidth]{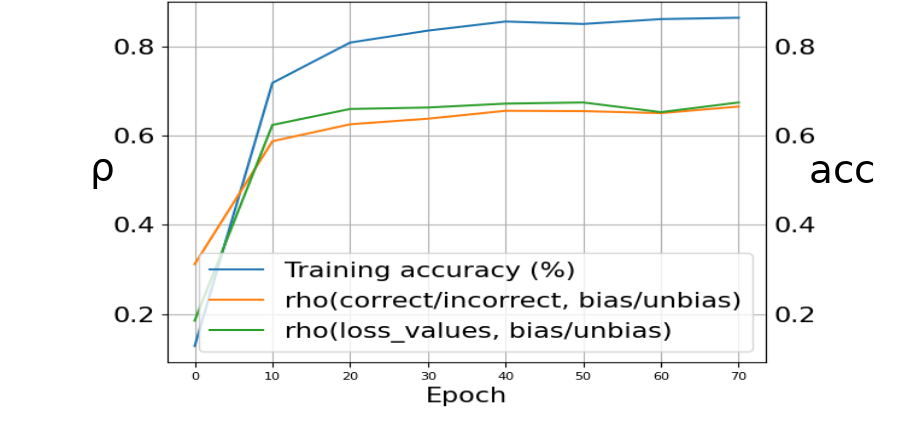}
    \caption{\footnotesize Training accuracy and Pearson correlation coefficient $\rho$ evolution over time. $\rho$ is computed between ground truth labels for $\mathcal{D}_{bias}$, $\mathcal{D}_{unbias}$ and the binary correct/incorrect vector $\mathbf{s}$. Similarly we compute $\rho$ for bias/unbias and the loss values per sample.
    }
    \label{fig:correlation_bias}

\end{figure}

This behaviour can be observed by looking at the loss function trends of the two subsets (see Fig. \ref{fig:problem_description}(c)). 
As a proof of concept, we computed the Pearson Correlation Coefficient $\rho$ between correct/incorrect predictions and bias/unbias ground-truth labels for 
CIFAR-10 at different epochs of training of an ERM model (see Fig. \ref{fig:correlation_bias}, orange plot). We observe that $\rho$ is sensibly higher than 0  as soon as the model fits the training data sufficiently (around $\rho=0.6$ when the training accuracy reaches $75\%$). We computed the same metric for the loss value with similar results (green plot). This shows that predictions of a vanilla ERM model are sufficiently correlated with ground truth bias/unbias labels, and therefore they can be exploited to divide the dataset.

We propose two approaches to assign to each samples the \textit{bias}/\textit{unbias} pseudo-label: the first one relies on predictions of an ERM model estimated at a specific epoch of the training process, where the epoch is determined fixing an achievable target training accuracy. The second approach instead takes into account the entire history of the predictions of a model along all of the training epochs. We call it identification ``by single prediction'' (SP) and ``by prediction history'' (PH), respectively. 

\subsection{Identification by single prediction}

The SP stage consists in training a neural network $f_{\phi}$, with parameters $\phi$, via ERM until it reaches a target training accuracy of $\gamma$, where $\gamma$ is a hyper-parameter. When the model reaches the desired accuracy level, the training stops and a forward pass of the entire training set is performed: samples that are correctly predicted are assigned to $\hat{\mathcal{D}}_{bias}$ while those not correctly predicted are assigned to $\hat{\mathcal{D}}_{unbias}$. This should capture the fact that biased examples are fitted more quickly by the model.
More formally:

\begin{equation}
\begin{aligned}
  \hat{\mathcal{D}}_{bias}^\gamma   = \{(x,y) \in \mathcal{D} \mid  \sigma ( f_{\phi}^\gamma(x) ) = y  \} \\
  \hat{\mathcal{D}}_{unbias}^\gamma = \{(x,y) \in \mathcal{D} \mid  \sigma ( f_{\phi}^\gamma(x) ) \neq y  \}
\end{aligned}
\label{eq:pseudo_labeling}
\end{equation}

Using $\gamma$ as a hyper-parameter is convenient for two reasons. 
First, the setting of the amount of desired accuracy is dataset agnostic. This is different from prior work \cite{JustTrainTwice} that employs a similar strategy, but with the hyper-parameter controlling the number of epochs to train the model: in such a case, the number of epochs are strictly dependent on the dataset the model is trained on.
Second, we can have a precise control of the amount of samples  assigned to the two splits, e.g. $\gamma = 0.85$ implies that 85\% of training data are assigned to $\hat{\mathcal{D}}_{bias}$ and 15\% to $\hat{\mathcal{D}}_{unbias}$. In real use cases, we do not know the correct assignments of the samples to the splits, so we have to rely on a a-priori setting of this parameter. When dealing with biased dataset, we consider the case in which high percentages of samples are biased, e.g. 95\% as in previous works \cite{nam2020learning, biaswap}: therefore setting $\gamma$ to high values is hardly a problem since the model overfits easily the training set given the multitude of biased samples.

\subsection{Identification by prediction history}
The PH method, instead of looking only at predictions at a specific epoch, considers the history of samples' predictions throughout the entire training stage. 
The rationale here is to make the splitting process more robust given that easy (biased) samples are often predicted correctly from the early training stages (often throughout the entire training), whereas hard (unbiased) samples could be difficult to be fit or are even left unlearned by the model. A second reason to prefer a multiple prediction approach is the fact that a prediction from a single epoch may be not representative of the actual difficulty of a specific sample: grouping predictions together is less affected by statistical oscillations.

Hence, the PH procedure consists in ERM training as for the SP method, but at the end of each epoch, we perform a forward pass of the entire training set and check whether the prediction is correct or not, building a binary vector having the dimension of the training set $N$. 
We then define the variable $s^t_i$, which is equal to $0$ if sample $i$ has been incorrectly classified at epoch $t$, or equal to $1$ vice versa. Hence, we can compute a vector $\mathbf{s^t}= \{s^t_i \}_{i=1}^N$ of dimension $N \times 1$, that indicates what samples have been correctly/incorrectly classified at epoch $t$. At the end of training, we will have the $N \times K$ matrix $\mathbf{S}$, with $K$ total number of training epochs, which  describes the history of predictions for each sample.

\begin{figure}[h!]
    \centering
    \includegraphics[width=0.8\textwidth]{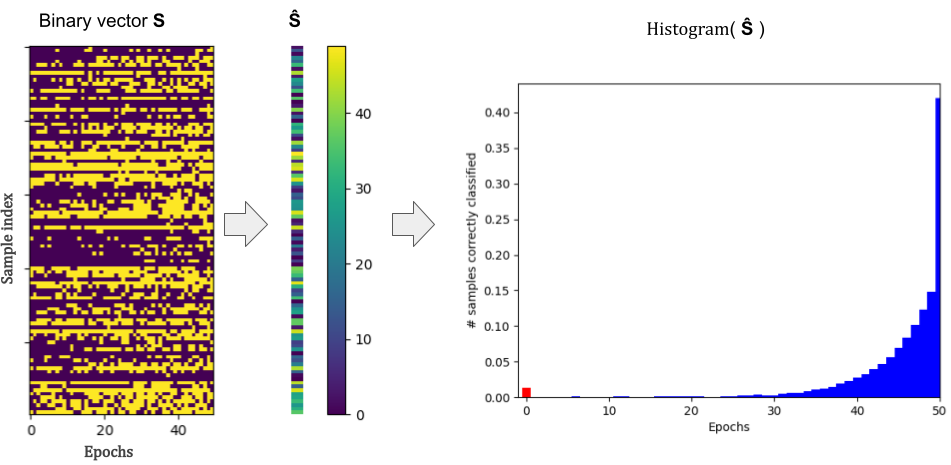}
    \caption{Binary vector $\mathbf{S}$ on the left: rows represent samples indexes $i$ while columns represent epochs $t$. $s_i^t$ is $0$ when sample $i$ is misclassified at epoch $t$ or $1$ when it is correctly classified. When summing up along the epochs, we get the vector $\hat{s} = \sum_{t=1}^{K} s^t$ (middle) that represent the ``prediction's history'' of each sample after $K$ epochs. When computing its histogram (right), we observe that the distribution is severely skewed towards easy samples (those correctly predicted most of the epochs). We highlighted in red the leftmost bin which is significantly higher than its neighbors: this shows how samples that are never correctly classified are more than those that are rarely correctly classified.} 
    \label{fig:ranking_BI}
\end{figure}

Summing up along the epoch axis produces an $N \times 1$ ranking vector $\hat{s}$ that describes how many epochs each sample has been correctly classified (see Figure \ref{fig:ranking_BI}, left). 
When computing the histogram for $\hat{s}$ (using $K$ bins - in order to represent all possible values of the entries of $\hat{s}$), one observes that most of the samples are correctly predicted throughout the entire training ($K$\textit{-th} rightmost bin in Fig. \ref{fig:ranking_BI}), whereas few samples are never or almost never correctly predicted (leftmost bin in Fig. \ref{fig:ranking_BI}, right).
Empirically, we found that most of the samples in the leftmost bins are unbiased. Provided with this insight, we devised a strategy to iteratively split the dataset by identifying the unbiased samples that are assigned to $\hat{\mathcal{D}}_{unbias}$, while assigning the others to $\hat{\mathcal{D}}_{bias}$.

We train an ERM model with cross entropy loss weighted by $\textbf{c}=\{c_i\}_{i=1}^N$, where $c_i$ is the weight for sample $i$, assuming values in $[0,1]$. Values equal or close to $1$ keep the learning rate unaltered or almost unaltered, whereas smaller values impact the learning rate by slowing it down. 
In fact, we now would like a model that enhances the differences between bias/unbiased samples, for better discriminating them: the vector $\textbf{c}$ is thus initialized with all entries equal to $1$ and then are progressively modified for this purpose. During training, we leave unaltered the weights for samples that are already well predicted while we decrease them for samples that are difficult to predict, thus implicitly decreasing their learning rate. Every $M$ epochs we compute the ranking vector $\sum_{t=1}^{M} s^t$ and use it to modulate the weights vector $\mathbf{c}$ as follows: 

\begin{equation}
    c^{new}_i = c^{old}_i - (1 - \frac{\sum_{t=1}^{M} s^t_i + 1}{M}).
    \label{eq:BI_coef}
\end{equation}

The value of $c_i$ is clipped in the range $[0,1]$, so that the learning rate can only be decreased or left unaltered. In this way, samples that are inherently difficult to be learned are even slowed down by this mechanism. 

When computing the histogram of the ranking vector at the end of training, we select all samples in the most difficult - leftmost - bin (those that have never been correctly predicted - see Fig. \ref{fig:ranking_BI}, right) and assign them to $\hat{\mathcal{D}}_{unbias}$, while assigning the others to $\hat{\mathcal{D}}_{bias}$.

\section{Bias-invariant representation learning}
\label{sec:repr_learn}

\noindent Provided with pseudo-labels for the two estimated subsets $\hat{\mathcal{D}}_{bias}$ and $\hat{\mathcal{D}}_{unbias}$, we now have to deal with the problem of learning data representations that are not only suitable for the biased data, but can generalize well to unbiased samples too. The driving idea is to exploit the two resulting subsets, even if noisy, to generate augmented data in order to counterbalance the harmful effect of the numerous biased examples in training. Hence, we propose to combine biased and unbiased samples as an effective strategy to produce synthetic images that exhibit attributes from both $\hat{\mathcal{D}}_{bias}$ and $\hat{\mathcal{D}}_{unbias}$.

As a preliminary analysis, we show in Sec. \ref{subsec:s-mix} how a convex combination \cite{mixup} of biased/unbiased samples results in a strong regularization, that helps preventing the model from overfitting the biased data, when added to standard weighted average of the loss terms for the two noisy subsets \cite{Ragonesi_2023_CVPR}. We will refer to this baseline as \textit{simple mix} or, in short, \textit{s-mix}. Motivated by the fact that the mixing coefficients have a strong impact on the final accuracy, we thus explore in Section \ref{subsec:l-mix} how the strategy for mixing the samples can be \textit{learned} directly from the data itself, in order to bring further improvement, leading to the proposed \textit{learnable mix} or \textit{l-mix} in short.

\begin{figure*}[!t]

    \centering
    \includegraphics[width=0.65\textwidth]{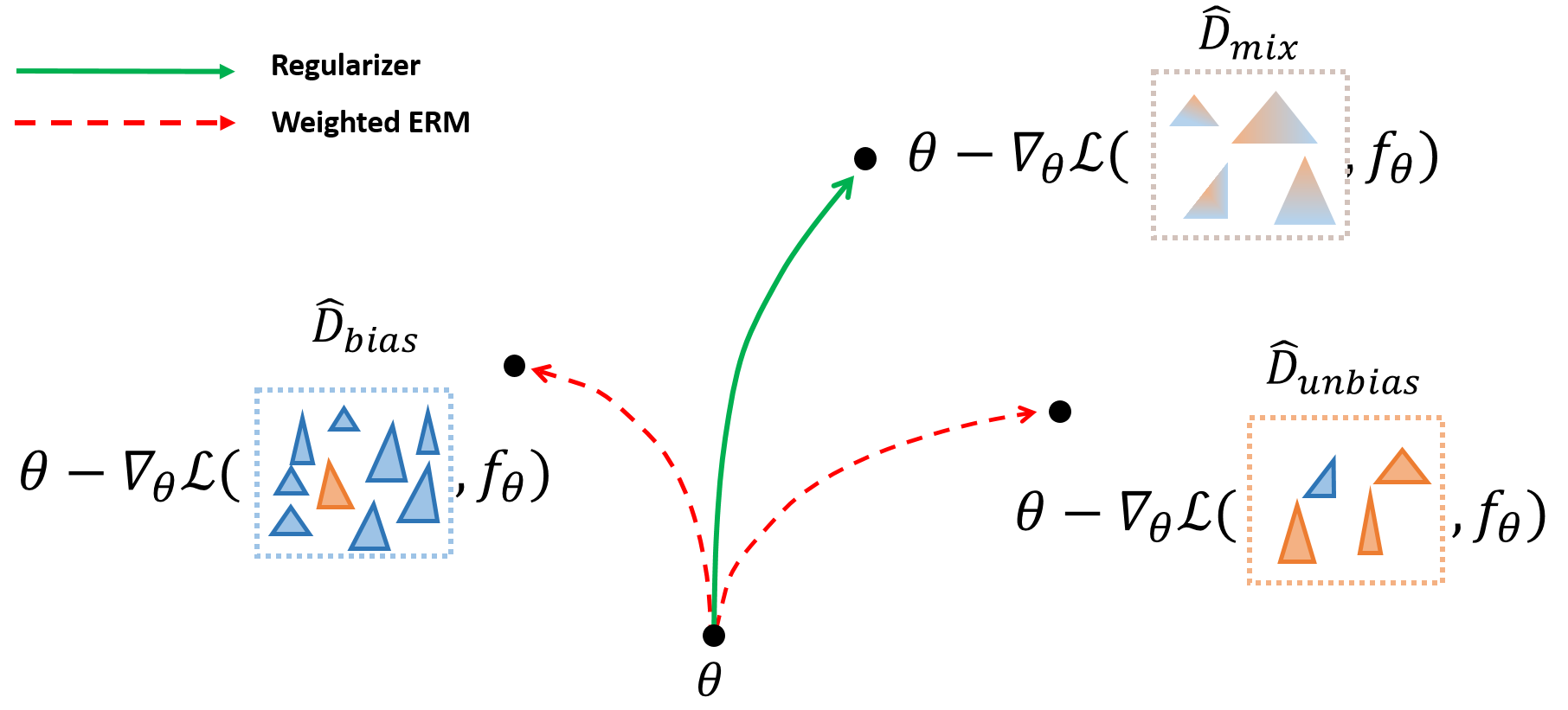}
    \caption{Gradients for $\mathcal{L} (\hat{\mathcal{D}}_{bias}, f_{\theta})$ and $\mathcal{L} (\hat{\mathcal{D}}_{unbias}, f_{\theta})$ are evaluated to produce the weighted ERM contribution. The regularization step using mixed data aims at producing an additional gradient contribution that decreases the loss function for $\hat{\mathcal{D}}_{mix}$ as for Equation \ref{eq:loss_method1}. (Best viewed in color) }
    \label{fig:mixup}

\end{figure*}

\subsection{Vanilla Mixup as a bias regularizer} 
\label{subsec:s-mix}

Given the two estimated pseudo-labeled subsets $\hat{\mathcal{D}}_{bias}$ and $\hat{\mathcal{D}}_{unbias}$, we  consider a neural network $f_{\theta}$, with parameters $\theta$, trained from scratch.

\vspace{0.2cm}
\noindent \textbf{Weighted ERM.} In this step we seek the best parameters $\theta$ on the basis of the two subsets $\hat{\mathcal{D}}_{bias}$ and $\hat{\mathcal{D}}_{unbias}$ via gradient descent. 
The $\theta$'s updating rule is:

\begin{equation}
    \theta^*=\theta - \eta ~\nabla_{\theta} ~[ 
    (1-\gamma) ~ \mathcal{L} (\hat{\mathcal{D}}_{bias}, f_{\theta}) +
    \gamma ~ \mathcal{L} (\hat{\mathcal{D}}_{unbias}, f_{\theta}) ]
    \label{theta_star}
\end{equation}

\noindent where $\eta$ is the learning rate. Note that in case the two subsets are obtained by \textit{single prediction}, $\gamma$ is the same hyper-parameter introduced in the Sect. 4.1. In this step, we scale the two loss functions with two coefficients $\gamma$ and $1-\gamma$ to deal with data imbalance ($|\hat{\mathcal{D}}_{bias}| >> |\hat{\mathcal{D}}_{unbias}|$). To rebalance the contributions from the two splits, an obvious choice is to set the weights inversely proportional to the cardinality of the two subsets, which is nothing else than the fixed and controllable hyper-parameter $\gamma$ used to estimate the biased and unbiased subgroups.

\vspace{0.2cm}
\noindent \textbf{Mixup as a regularizer.} Subsequently, we seek a representation that can conciliate both biased and unbiased samples and at the same time can prevent the model from overfitting the training data (especially $\hat{\mathcal{D}}_{bias}$).
We take inspiration from Mixup \cite{mixup} as a way to combine samples from the two subsets. Mixup consists in using a convex combination of both input samples and labels demonstrating its efficacy as an effective regularizer. 
Specifically, we feed the model with synthetic (augmented) samples resulting from the \textit{mix of examples from biased and unbiased subsets}, aiming at breaking the shortcuts present in the data (see Fig. \ref{fig:mixup}).

\noindent We construct $\hat{\mathcal{D}}_{mix}=\{x_{mix},y_{mix}\}$ by mixing examples of $\hat{\mathcal{D}}_{bias}$, $\hat{\mathcal{D}}_{unbias}$ with corresponding labels $y$, as in the Mixup method \cite{mixup}, by sampling the mixing parameter $\lambda$ from a Beta distribution, i.e.:

\begin{equation}
\begin{aligned}
  (\hat{x}_1, \hat{y}_1) \in \hat{\mathcal{D}}_{bias}~,  (\hat{x}_2, \hat{y}_2) \in \hat{\mathcal{D}}_{unbias} \\
  x_{mix} = \lambda ~ \hat{x}_1 + (1-\lambda) ~ \hat{x}_2 \\
  y_{mix} = \lambda ~ \hat{y}_1 + (1-\lambda) ~ \hat{y}_2 \\
  with  \quad  \lambda \sim Beta(\alpha, \beta) \\
\end{aligned}
\label{eq:mixup}
\end{equation}

Once the augmented samples ($x_{mix},y_{mix}$) are computed, the model is updated combining the weighted ERM and the regularization term:
\begin{equation}
\small
\begin{aligned}
    \mathcal{L} := 
    \underbrace{
    (1-\gamma) ~ \mathcal{L} (\hat{\mathcal{D}}_{bias},   f_{\theta})  +
    \gamma ~ \mathcal{L} (\hat{\mathcal{D}}_{unbias}, f_{\theta}) }_{\text{Weighted ERM}}   +~\zeta ~ \underbrace{ \mathcal{L} ( \hat{\mathcal{D}}_{mix}, f_{\theta}) }_{\text{Regularizer}} 
\label{eq:loss_method1}
\end{aligned}
\end{equation}

\noindent where $\zeta$ is a hyper-parameter weighting the regularization term. 

The hyper-parameter $\zeta$ controls the amount of regularization in the final loss: if $\zeta=0$, the method corresponds to standard (weighted) ERM in which the contributions of the losses on the two subsets are scaled by $(1-\gamma)$ and $\gamma$. When $\zeta>0$ the weighted ERM optimization trajectory is corrected by the regularization term (Fig. \ref{fig:mixup}). This corresponds to find parameters $\theta$ that are good for both $\hat{\mathcal{D}}_{bias}$ and $\hat{\mathcal{D}}_{unbias}$, but can also eventually reduce the loss value on the newly generated data samples $\hat{\mathcal{D}}_{mix}$. 
We will actually show that performance is not much affected by the choice of $\zeta$: as detailed in Sect. \ref{subsec-zetaablation}, accuracy increases and reaches a plateau, and only with higher orders of magnitude  (in the order of $\zeta\sim 10^3$) the regularization effect becomes detrimental. We set $\zeta=10$ for all experiments.

\begin{figure*}[t]
    \centering
    \includegraphics[width=\textwidth]{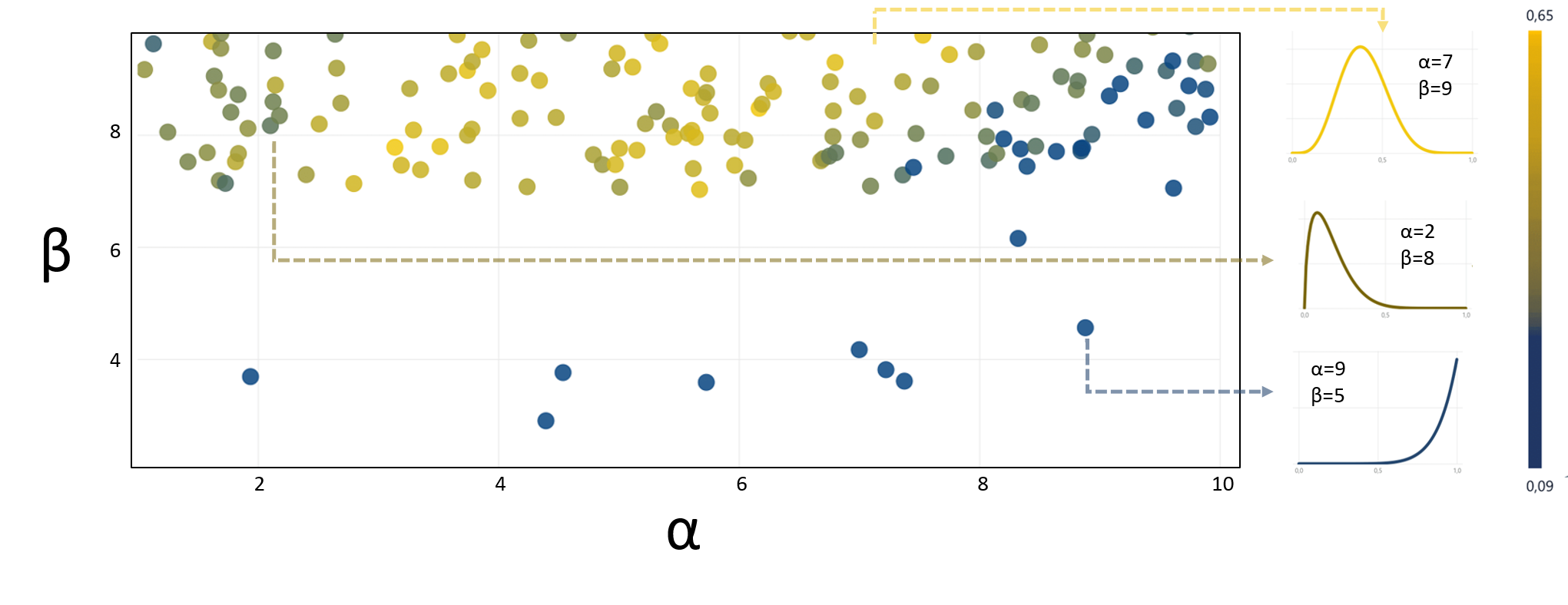}
    \caption{Performance of a model trained with augmented samples generated by Mixup for different pairs of parameters $(\alpha, \beta)$ (equations \ref{eq:mixup} and \ref{eq:loss_method1}). Each dot represents a single run done with parameters of corresponding coordinates. The right color bar represents the model's test accuracy. We show 3 examples for different probability density functions corresponding to specific pairs $(\alpha, \beta)$.}
    \label{fig:beta_distrib_chart}

\end{figure*}

To conclude this preliminary analysis, we explored how the choice of the two parameters $(\alpha, \beta)$ impacts the final test accuracy on a standard dataset, CIFAR-10.  As shown in Fig. \ref{fig:beta_distrib_chart} the convex combination of biased and unbiased samples works reasonably well for a wide range of $\alpha$ and $\beta$ that control the probability density function of the Beta distribution (in this preliminary investigation only, we used the ground truth annotations for $\mathcal{D}_{bias}$ and $\mathcal{D}_{unbias}$ in order to decouple the learning problem from the bias-identification problem). We employed Bayesian optimization to explore the parameters space in order to highlight the region in which the generated samples gives the best accuracy on the test set. Figure \ref{fig:beta_distrib_chart} shows that the best combination is attained when the Beta distribution is slightly skewed towards the unbiased samples, namely, they are given more importance than the biased ones. Vice versa, the worst scenario occurs when the distribution puts more emphasis to the biased samples. However, we claim that augmented data can be generated in a more principled way, that is, by \textit{learning} $\alpha$ and $\beta$ as detailed in the next section.

\subsection{Unbiasing by learnable data augmentation}
\label{subsec:l-mix}

\begin{figure*}[!t]
    \centering
    \includegraphics[width=0.85\textwidth]{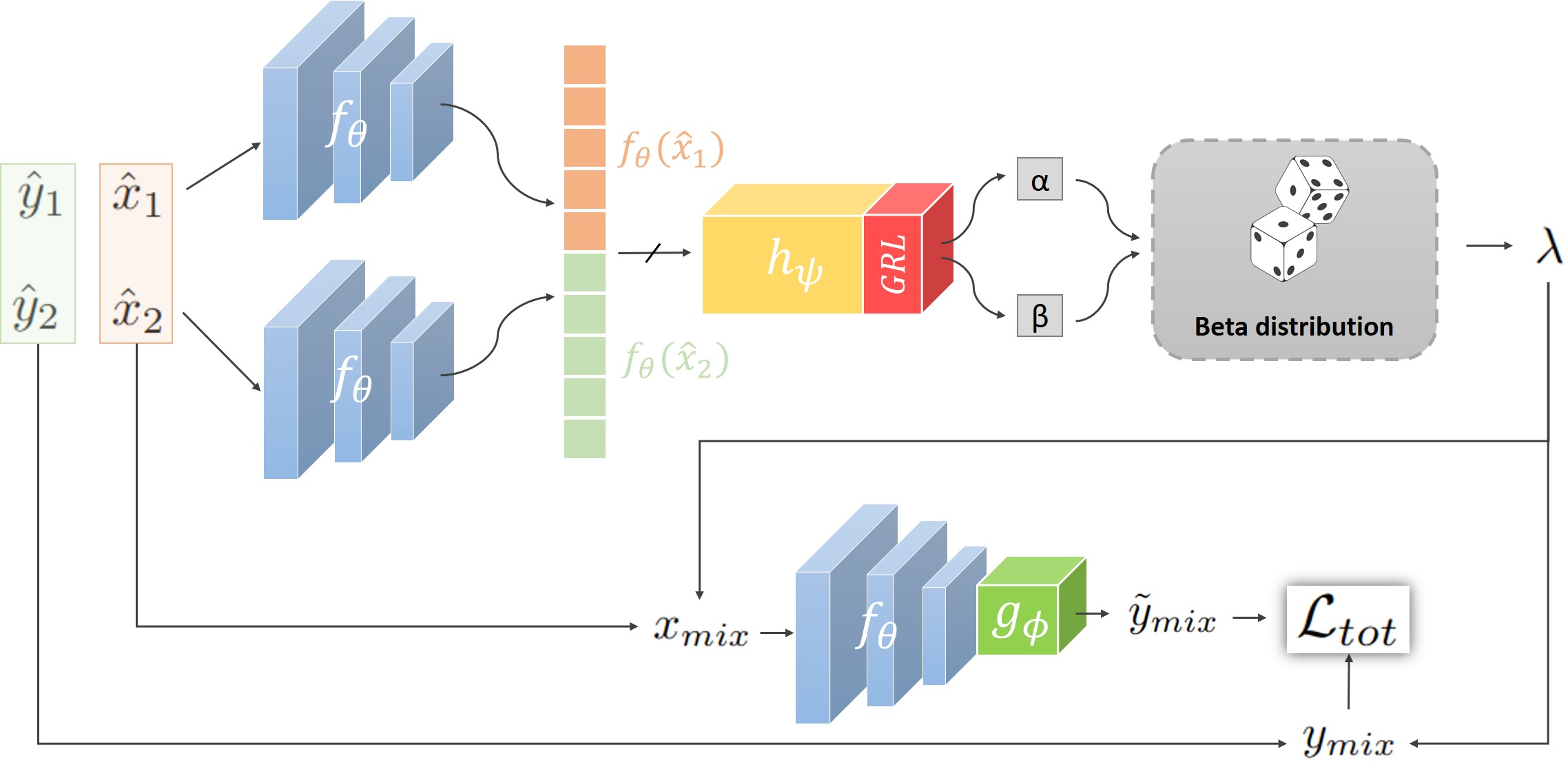}
    \caption{Proposed pipeline for learning the parameters of the Beta distribution. Features from $(\hat{x}_1, \hat{y}_1)\in \hat{\mathcal{D}}_{bias}$  and $(\hat{x}_2, \hat{y}_2)  \in \hat{\mathcal{D}}_{unbias}$ are extracted with $f_{\theta}$, concatenated and fed to the module $h_{\psi}$, which predicts the $\alpha$ and $\beta$ that parameterize a Beta distribution, which, in turn, is sampled to obtain the mixing coefficient $\lambda$. The GRL layer on top of $h_{\psi}$ model ensures that when gradients are back-propagated, $\psi$ parameters are optimized to \textit{maximize} the cross entropy. Namely, we want to produce mixed samples that are challenging for the classification network $g_{\phi}(f_{\theta})$. Note also that gradients are stopped before going back to the initial encoders ($\nrightarrow$ denotes a StopGradient operation).}
    \label{fig:gradient}

\end{figure*}

Building on the insights gained from the analysis of Sec. \ref{subsec:s-mix}, we designed an end-to-end pipeline that is able to simultaneously learn the classification task and optimize for the best mixing strategy. 
Our model (see Fig. \ref{fig:gradient}) is composed of 1) a backbone network acting as feature extractor $f_{\theta}$, and 2) a classifier $g_{\phi}$, with parameters $\theta$ and $\phi$, respectively, the latter producing logits from the features $f_{\theta}(x)$.We also adopt 3) a module $h_{\psi}$ which outputs two scalar values, namely, the parameters $\alpha$ and $\beta$ of the Beta distribution. Samples from $\hat{\mathcal{D}}_{bias}$ and $\hat{\mathcal{D}}_{unbias}$ are fed to $f_{\theta}$ and the corresponding feature vectors $f_{\theta}(\hat{x}_1)$ and $f_{\theta}(\hat{x}_2)$ are then concatenated. The module $h_{\psi}$ takes as input such concatenation, denoted as $\mathbin\Vert$ in the equation below, and estimates the parameters $(\alpha, \beta)$ of the Beta distribution, which is subsequently sampled to extract the mixing coefficient $\lambda$:
\begin{equation}
\label{eq:alphabeta}
  (\alpha, \beta) = h_{\psi}(\, f_{\theta}(\hat{x}_1) \mathbin\Vert f_{\theta}(\hat{x}_2)\,)  
\end{equation}

Specifically, we sample a vector of mixing coefficients $\lambda$'s (one for each pair in the batch) from the Beta distribution.  We rely on the reparametrization trick \cite{vae} in order to have a fully differentiable pipeline, since we need to differentiate through the sampling procedure. To do so, we employed the Pytorch implementation of the reparametrized Beta distribution. 
More formally, we sample $\lambda_i$ for the $i-$sample from the Beta distribution parametrized by $\alpha^*_i$ and $\beta^*_i$: \begin{equation}
    \lambda_i \sim Beta(\alpha^*_i, \beta^*_i).
\end{equation}
Provided with $\boldsymbol{\lambda}=\{\lambda_i\}_{i=0}^N$, we compute $(x_{mix}, y_{mix})$ (Eq. \ref{eq:mixup}), and use it as input for $f_{\theta}$, whose outcome is subsequently fed to $g_{\phi}$.
The logits are then passed to the $g_{\phi}$ softmax layer $\sigma$ to infer $\tilde{y}_{mix}$, which is used as input for the Cross Entropy (CE) loss $\mathcal{L}_{CE}$. \\

Following the paradigm of adversarial training, we want to learn combinations of $x_1$, $x_2$ that generate challenging samples for the network, acting indeed as adversaries, i.e., for which the loss value \textit{increases}. Such augmented samples are then used as training samples for effectively minimizing the loss function.

When doing backpropagation we want to seek the parameters $\theta$ and $\phi$ that minimize the CE loss, while at the same time looking for the combination -- i.e., the parameter $\lambda$ drawn from the Beta distribution governed by parameters $\alpha$ and $\beta$ -- of the two inputs $\hat{x}_1$ and $\hat{x}_2$ that maximize the CE loss. To do so, we apply a Gradient Reversal Layer (GRL) \cite{grl} to change the sign of the gradient when updating the weights ${\psi}$ (see Fig. \ref{fig:gradient}). \\

Formally, the optimization problem we want to solve is:

\begin{equation}
\begin{aligned}
    \theta^*, \phi^* = \underset{\theta, \phi}{\mathrm{argmin}} ~~\mathcal{L}_{CE}(\sigma(g_{\phi}(f_{\theta}(x_{mix})), y_{mix})\\
    \psi^* = \underset{\psi}{\mathrm{argmax}} ~~\mathcal{L}_{CE}(\sigma(g_{\phi}(f_{\theta}(x_{mix})), y_{mix})\\
\end{aligned}
\label{eq:optimi}
\end{equation}

In the second equation the dependence on $\psi$ is implicit in the variables $(x_{mix}, y_{mix})$ as defined in equation \ref{eq:mixup}, since the mixing coefficients $\lambda_i$ are sampled from $Beta(\alpha_i, \beta_i)$, where $\alpha$ and $\beta$ depend upon $\psi$ as from equation \ref{eq:alphabeta}.
Note that we prevent the gradient to flow into $f_{\theta}$ for the two streams (Fig. \ref{fig:gradient}). Therefore, $f_{\theta}$ is updated only through $\frac{\partial f_\theta(x_{mix})}{\partial \theta}$. 

\begin{algorithm}[H]
    \caption{Pseudocode for a single training iteration.}
    \begin{algorithmic}[1]
        \Function{training\_iteration}{$\hat{\mathcal{D}}_{bias}, \hat{\mathcal{D}}_{unbias}, \eta$}\\
            \State sample batches $(\hat{x}_1, \hat{y}_1), (\hat{x}_2, \hat{y}_2)$ from $\hat{\mathcal{D}}_{bias}, \hat{\mathcal{D}}_{unbias}$
            \State compute $f_{\theta}$($\hat{x}_1$),$f_{\theta}$($\hat{x}_2$)

            \State detach gradient from $f_{\theta}$($\hat{x}_1$), $f_{\theta}$($\hat{x}_2$)

            \State $(\alpha, \beta) = h_{\psi}(\, f_{\theta}(\hat{x_1}) \mathbin\Vert f_{\theta}(\hat{x_2})\,)$\\
            
            \State $\lambda$ $\gets$ sample from Beta($\alpha$, $\beta$)
            
            \State $x_{mix} = \lambda ~ \hat{x}_1 + (1-\lambda) ~ \hat{x}_2 $
            \State $y_{mix} = \lambda ~ \hat{y}_1 + (1-\lambda) ~ \hat{y}_2 $\\
            
            \State $\Tilde{y}_{mix} \gets g_{\phi}(f_{\theta}(x_{mix}))$
            \State $\mathcal{L}_{tot} \gets \mathcal{L}_{CE}(\Tilde{y}_{mix}, y_{mix}) + \omega Reg$ \\

            \State{\# update networks' parameters}
            \State $\theta \gets \theta - \eta\nabla_{\theta}\mathcal{L}_{tot}$
            \State $\phi \gets \phi - \eta\nabla_{\phi}\mathcal{L}_{tot}$
            \State $\psi \gets \psi + \eta\nabla_{\psi}\mathcal{L}_{tot}$ \Comment{GRL inverts gradient sign}
            
        \EndFunction
    \end{algorithmic}
\end{algorithm}

\vspace{0.2cm}
\noindent \textbf{Regularizer.} We may optionally introduce a regularizer that acts as a prior knowledge on the objective function to lead to skewed distributions. Since we deal with imbalanced data, i.e. $| \mathcal{D}_{bias} | >>  | \mathcal{D}_{unbias} | $, it is usually beneficial (see Fig. \ref{fig:beta_distrib_chart}) to constrain $h_{\psi}$ to learn a family of Beta distributions that are skewed towards the minority group, i.e., the unbiased subset. This can be easily achieved by adding a regularization term $Reg$ that forces the expected value of the Beta distribution to be close to the (noisy) bias ratio. We chose the simple Mean Square Error as regularizer: 

\begin{equation}
    Reg = MSE(~\frac{\alpha}{\alpha + \beta},~ \frac{| \hat{\mathcal{D}}_{unbias} |}{| \hat{\mathcal{D}}_{bias} | + | \hat{\mathcal{D}}_{unbias} |}~).
\end{equation}

Thus, the total loss becomes:

\begin{equation}
    \mathcal{L}_{tot} = \mathcal{L}_{CE}(\sigma(g_{\phi}(f_{\theta}(x_{mix})), y_{mix}) +~\omega ~Reg, 
    \label{eq:loss_method2}
\end{equation}
where $\omega$ is a hyper-parameter that controls the amount of the regularization. 

The several stages composing the proposed method and all introduced hyper-parameters will be analyzed in our ablation study in Sect. 7.
\\

\section{Experiments}
\label{sec:experiments}

We tested our proposed method on different benchmarks, ranging from datasets with controlled synthetic biases to more realistic image classification tasks. We made a comparison with other existing methods tackling the debiasing problem in both supervised and, more extensively, unsupervised ways.

\subsection{Benchmarks}
\label{subsec:benchmarks}
\noindent \textbf{Synthetic bias.} 
We consider a benchmark that has been introduced in \cite{corrupted_cifar}, namely corrupted CIFAR-10 in which the amount of bias is synthetically injected and  thus controlled. 
Corrupted CIFAR-10 is a modification of the original dataset \cite{cifar10} and has been adopted by Nam et al. \cite{nam2020learning} in the context of debiasing. It consists of 50,000 training RGB images and 10 classes. The bias stems from the fact that each image is corrupted with a specific noise/effect (e.g., motion blur, shot noise, contrast enhancement, etc.). In practice, each class has a privileged type of noise under which it is observed during training (e.g., most of truck images are corrupted with Gaussian blur). 

\vspace{0.3cm}
\noindent \textbf{Realistic bias.} 
We prove the efficacy of our method on realistic image datasets, namely, Waterbirds, CelebA and Bias Action Recognition (BAR). 
Waterbirds has been introduced in \cite{group_DRO} and combines bird photos from the Caltech-UCSD Birds-200-2011 (CUB) dataset \cite{CUB_dataset}
where the background is replaced with images from the Places dataset \cite{places_dataset}. It consists of 4,795 training images and the goal is to distinguish two classes, namely \textit{landbird} and \textit{waterbird}. The bias is represented by the background of the images: most landbirds are observed on a land background while most waterbirds are observed in a marine environment. 

CelebA \cite{celebA_dataset} consists in over 200,000 celebrity face photos annotated with 40 binary attributes (e.g. smiling, mustache, brown hair, etc.). We consider a subset of 162,770 photos and solve the task of deciding, whether the image depicts a blonde person or not: this is the same setting considered in past works \cite{group_DRO, JustTrainTwice} that tackle the debiasing problem. The blonde/non blonde binary label is spuriously correlated with the gender label (the minority group blond-male contains only 1387 images). 

BAR has been introduced in \cite{nam2020learning} as a realistic benchmark to test model's debiasing capabilities. It is constructed using several data sources and contains 1,941 photos of people performing several actions. The task is to distinguish between 6 categories: Climbing, Diving, Fishing, Racing, Throwing and Vaulting. The bias arises from the context in which action photos are observed at training time: for instance, diving actions are performed in a natural environment employing a diving suit at training time, whereas in the test set, they are set in an artificial environment, e.g. a swimming pool. For details, readers can refer to the original paper \cite{nam2020learning}.

\begin{figure*}[t!]
    \centering
    \includegraphics[width=0.95\textwidth]{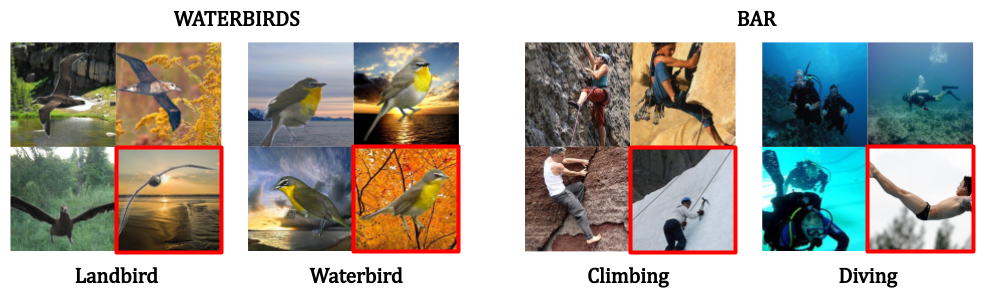}
    \caption{ \footnotesize Examples of training biased and unbiased (with red boundary) data, from Waterbirds and BAR.}

    \label{fig:model}
    
\end{figure*}

\subsection{Performances}
\noindent We report the performance of our approach on the different benchmarks above mentioned. Accuracy is the metric adopted to evaluate the performance: since we deal with biased training data and balanced data in testing, we are primarily interested in improving accuracy on the unbiased test samples, those under-represented in the training data. However, at the same time, we do not want to lose performance on the biased samples as we learn more general features: in fact, a higher generalization implies that spurious correlations are not used anymore for classification, and this may cause a drop in performance on such samples. For this reason, we report accuracy on the testing subset of unbiased samples only, as well as over the entire test set (biased $+$ unbiased).

\vspace{0.3cm}
\noindent {\bf Implementation details.} 
All experiments comply the same evaluation protocol used in the competing methods for a fair comparison. 
We used ResNet-18 \cite{resnet} as a backbone for Corrupted CIFAR-10 and BAR, and ResNet-50 as backbone for Waterbirds and CelebA. We remove the last layer from such backbones, adding a 2-layer MLP head on top of it as a classifier. ResNets are pre-trained on ImageNet \cite{imagenet}. 
The parameter network $h_{\psi}$ consists of a simple MLP with two hidden layers with $64$ neurons each. We rely on the Pytorch implementation of the Beta distribution for sampling in a reparameterized fashion. We set the learning rate $\eta=0.001$ for all datasets with batch size of $256$ on synthetic biased data and $128$ for realistic bias data. We used Adam \cite{adam_optimizer} as optimizer. 
In all experiments we adopted the prediction history (PH) method to infer the biased/unbiased samples. In our ablation analysis in Sec. \ref{sec:ablation}, we discuss how different splitting strategies influence the final outcome, as well as further implementation details.

\begin{table*}[!ht]
\resizebox{\textwidth}{!}{

\begin{tabular}{c|c|cc|cccc}

\toprule

\textbf{Bias} & & \multicolumn{2}{c|}{\textbf{Supervised}} & \multicolumn{4}{c}{\textbf{Unsupervised}} \\

\textbf{ratio} & \textbf{ERM} &  \textbf{REPAIR \cite{li2019repair}} & \textbf{G-DRO \cite{group_DRO}} & \textbf{LfF \cite{nam2020learning}} & \textbf{SelecMix \cite{Hwang22}} & \textbf{BiaSwap \cite{biaswap}}  & \textbf{Ours
} \\

\midrule

$95\%$ & $45.2 \pm 0.22$ & $48.7$ & $53.1$ & $59.9$ & $54.00$ & $46.99$  & \textbf{64.7 $\pm $ 1.20}\\

 $98\%$ & $30.2 \pm 0.77$ & $37.9$ & $40.2$ & $49.4$ & $47.70$ & $41.16$  & \textbf{57.4 $\pm $ 1.15}\\

$99\%$ & $22.7 \pm 0.97$ & $32.4$ & $32.1$ & $41.4$ & $41.87$ & $38.94$ & \textbf{50.8 $\pm $ 1.03}\\

$99.5\%$ & $17.9 \pm 0.86$ & $26.3$ & $29.3$ & $31.7$ & $38.14$ & $35.87$ &  \textbf{43.9 $\pm $ 0.87}\\

\bottomrule
\end{tabular}}

\caption{ \textbf{Accuracy on the whole test set of Corrupted CIFAR-10.} Accuracy (in \%) evaluated on \textit{biased $+$ unbiased} test samples for different bias ratios. Results of REPAIR, Group-DRO and LfF are from \cite{nam2020learning}. The performance of our proposed methods are obtained with
$\omega=10^{-3}$.
Best performance in bold.
}
\label{table_toy_test}
\end{table*}

\begin{table*}[!ht]
\resizebox{\textwidth}{!}{
\begin{tabular}{c|c|cc|cccc}
\toprule

\textbf{Bias}& & \multicolumn{2}{c|}{\textbf{Supervised}} & \multicolumn{4}{c}{\textbf{Unsupervised}} \\

 \textbf{ratio} & \textbf{ERM} &  \textbf{REPAIR \cite{li2019repair}} & \textbf{G-DRO \cite{group_DRO}} & \textbf{LfF \cite{nam2020learning}} & \textbf{SelecMix \cite{Hwang22}} & \textbf{BiaSwap \cite{biaswap}} & \textbf{Ours} 
\\
\midrule

$95\%$ & $39.4 \pm 0.75$ & $50.0$ & $49.0$ & $59.6$ & $51.38$ & $42.61$  & \textbf{64.6 $\pm $ 1.14}\\

$98\%$ & $22.6 \pm 0.45$ & $38.9$ & $35.1$ & $48.7$ & $44.24$ & $35.25$& \textbf{57.2 $\pm $ 1.09}\\

$99\%$ & $14.2 \pm 0.91$ & $33.0$ & $28.0$ & $39.5$ & $35.89$ & $32.54$ & \textbf{50.6 $\pm $ 1.01}\\

$99.5\%$ & $10.5 \pm 0.28$ & $26.5$ & $24.4$ & $28.6$ & $31.32$ & $29.11$  & \textbf{43.7 $\pm $ 0.99}\\

\bottomrule
\end{tabular}}

\caption{ \textbf{Results on the unbiased test samples of Corrupted CIFAR-10.} Accuracy (in \%) evaluated \textit{only} on the unbiased samples for different bias ratios. Results of REPAIR, Group-DRO and LfF are from \cite{nam2020learning}. The performance of our proposed methods are obtained with 
$\omega=10^{-3}$.
Best performance in bold.
}
\label{table_toy_bias}
\end{table*}

\vspace{0.3cm}
\noindent {\bf Performance for synthetic bias.} Tables \ref{table_toy_test} and \ref{table_toy_bias} present the performances (best and second best) on synthetic biased datasets, reporting the average accuracy (mean $+$ standard deviation) for the whole test set and for unbiased samples only, respectively. 

We compare against a model trained by Empirical Risk Minimization (ERM) as baseline, and different former methods to learn unbiased representations, either using annotation for the bias or not. 
For the methods requiring explicit knowledge of the bias (i.e., supervised), we consider REPAIR \cite{li2019repair}, which does sample upweighting, and Group-DRO \cite{group_DRO}, which tackles the problem using robust optimization. We also compare our performance with that of Learning from Failure (LfF) \cite{nam2020learning}, BiaSwap \cite{biaswap} and SelectMix \cite{Hwang22}, which learn a debiased model without exploiting the labeling of the bias (i.e., unsupervised).

We consider different ratios of the bias (ranging from $95\%$ up to $99.5\%$), as in \cite{nam2020learning}. This ratio indicates the actual percentage of the dataset belonging to $\mathcal{D}_{bias}$ and $\mathcal{D}_{unbias}$, i.e., $|\mathcal{D}_{bias}|$/$(|\mathcal{D}_{unbias}|+|\mathcal{D}_{bias}|)$. 
The choice of the method to estimate $\hat{\mathcal{D}}_{bias}$ and $\hat{\mathcal{D}}_{unbias}$ has a direct impact on the performance. The hyperparameter of the PH method, controlling the way the original dataset is being divided, is $M$ (see Sect. \ref{sec:BI}.B), which is set to $M=5$ throughout all the experiments. 
For \textit{l-mix}, we set the hyperparameter $\omega$ weighting the regularization term to $\omega=10^{-3}$ as a results of a careful ablation (Sec. \ref{subsec-omegaablation}).

For Corrupted CIFAR-10, our \textit{l-mix} method always reaches the best accuracy.
We can note here that the gap with former methods, both supervised and unsupervised, is much higher, ranging overall between about 5\% and 20\% regarding the whole test set (min and max difference across bias ratios, Table \ref{table_toy_test}), and between about 5\% and 22\%, when testing on the biased samples only (Table \ref{table_toy_bias}). It is important to highlight that the results of our proposed methods are reported as mean $\pm$ standard deviation, obtained after 5 run of the algorithms, while the accuracies reported by SelecMix and BiaSwap are the result of a single (presumably the best) run. Further, taking into account the standard deviation, our accuracies are comparable with the one of the winning methods.

Our results show that the obtained performances are rather stable and consistent across datasets and bias ratios, as compared to previous approaches.

\begin{table*}[h!]
\resizebox{\textwidth}{!}{
\begin{tabular}{lr|l|llll|l}
\toprule

 & & & \multicolumn{4}{c|}{\textbf{Unsupervised}} & \multicolumn{1}{c}{\textbf{Supervised}}  \\

\textbf{Dataset} & \textbf{Test set} & 
\textbf{ERM} &  \textbf{CVar DRO \cite{CVaR_DRO}} & \textbf{LfF \cite{nam2020learning}}  & \textbf{JTT \cite{JustTrainTwice}}  & \textbf{Ours} &
\textbf{G-DRO \cite{group_DRO}}\\

\midrule

\multirow{ 2}{*}{Waterbirds}  

& 
\textit{All} & $97.3$ & \textbf{96.0} & $91.2$ & $93.3$ & \underline{$94.7$} & $93.7$\\
& 
\textit{Unbiased} & $72.6^{\;\;\textit{(24.7)}\updownarrow}$ & $75.9^{\;\;\textit{(20.1)}\updownarrow}$ & $78.0^{\;\;\textit{(13.2)}\updownarrow}$  & \underline{$86.7$}$^{\;\;\textit{(6.6)}\updownarrow}$ & $\textbf{87.7}^{\;\;\textit{(7.0)}\updownarrow}$ & $91.4^{\;\;\textit{(2.3)}\updownarrow}$ \\

\midrule

\multirow{ 2}{*}{CelebA}  

& 
\textit{All} & $95.6$ & $82.5$ & $85.1$ & \underline{$88.0$}  & \textbf{89.1} &$92.9$\\
& 
\textit{Unbiased} & $47.2^{\;\;\textit{(48.4)}\updownarrow}$ & $64.4^{\;\;\textit{(18.1)}\updownarrow}$ & $77.2^{\;\;\textit{(7.9)}\updownarrow}$ & \underline{$81.1$}$^{\;\;\textit{(6.9)}\updownarrow}$ & $\textbf{82.2}^{\;\;\textit{(6.9)}\updownarrow}$ & $88.9^{\;\;\textit{(4.0)}\updownarrow}$ \\

\midrule

BAR

& 
\textit{All} & $53.5$     & -   & \underline{$62.9$} & - & \textbf{64.8} & -\\

\bottomrule
\end{tabular}}

\caption{ \textbf{Results on realistic datasets.} Accuracy (in \%) evaluated on the test set composed by unbiased test samples (Unbiased) and the entire test set (All). The performance of our proposed methods are obtained with $\omega=10^{-3}$. \textbf{Best results} among unsupervised methods for each dataset are in bold, \underline{second-best} are underlined. The gap between \textit{All} and \textit{Unbiased} accuracy are also reported in brackets and indicated by $\updownarrow$. 
}

\label{table_real_data}
\end{table*}

\vspace{0.3cm}
\noindent \textbf{Performance on the realistic biased datasets.}  
In these trials, we still compare against the ERM baseline and Group DRO, as supervised method, and four unsupervised algorithms, LfF \cite{nam2020learning}, CVaR DRO \cite{CVaR_DRO} and JTT \cite{JustTrainTwice}, the same protocol previously reported. Performances are reported in Table \ref{table_real_data}. For these datasets, we remind that we do not have the full control of the bias ratios. Specifically, in BAR we do not know exactly the biased/unbiased samples and,  
differently from Corrupted CIFAR-10, which have a balanced test set, Waterbirds and CelebA test sets are also imbalanced. 
In these cases, it is also important not only to cope with unbiased samples, but also to maintain accuracy on biased data.
Consequently, it is paramount here to reach a good trade-off between generalizing to unbiased samples while keeping high performance on biased data as well. Hence, performances in Table \ref{table_real_data} are reported as accuracies over both the entire test set (\textit{All}) and the unbiased samples (\textit{Unbiased}) for Waterbirds and CelebA, and over the whole test set only (\textit{All}) for BAR. 

\begin{figure}
    \centering
    \includegraphics[width=0.5\linewidth]{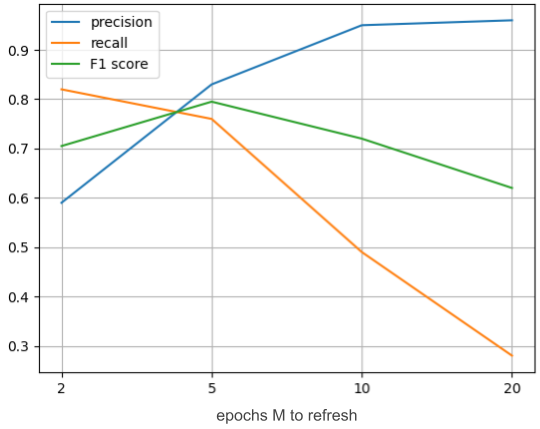} 

    \caption{Performance of the Bias Identification method on Corrupted CIFAR-10 (bias ratio $0.95$), for different parameter choice $M$, i.e. how often the coefficient vector is updated (Eq. \ref{eq:BI_coef}). We report Precision, Recall and F1-score as metrics. 
    }
    \label{fig:ablation_gamma}
\end{figure} 

For both Waterbirds and CelebA, we score favorably with respect to other unsupervised methods for the subset of \textit{unbiased} samples: we outperform the best state-of-the-art (JTT \cite{JustTrainTwice}) of about 1\%, while a larger gap is detected for other competing methods, namely, about 12\% and 10\% in Waterbirds, and 18\% and 5\% in CelebA, for CVar DRO  \cite{CVaR_DRO} and LfF \cite{nam2020learning}, respectively.

When addressing the entire test set (indicated as \textit{All} in the table, i.e. biased $+$ unbiased samples), we obtain the best accuracy for CelebA, outperforming JTT, LfF and CVar DRO by 1,1\%, 4\% and 6.6\%, respectively. For Waterbirds, our \textit{l-mix} is the second best scoring -1.3\% with respect to CVar DRO, followed by the other methods. In this respect, it is important to notice that CVar DRO, while it generally performs well on the entire, \textit{biased} test set, it drastically experiences large drops when considering the \textit{unbiased} samples only, equal to 20\% points and 18\% for Watebirds and CelebA, respectively. The same behavior is evident for the standard ERM, with even larger gaps, due to the fact that it is not paying attention to any data bias.

Overall, our proposed method reaches the best trade-off, resulting competitive for both the types of test sets, across the several datasets. In other words, we are able to learn bias invariant representations without giving up accuracy on the biased samples.

Finally, we show also competitive performance against the supervised method Group DRO: without using any bias supervision, on Waterbirds our method surpasses its test accuracy on the entire test set, even if the accuracy on biased data only results lower (owing to the supervision in this case). 
On CelebA, the performance are less competitive with Group DRO, yet we still outperform other unsupervised debiasing methods.  
Concerning the BAR dataset, since there is no ground-truth for the bias we report only the average accuracy over the whole test set: also inthis case, our method outperforms all other competitors by a considerable margin.

\section{Ablation analysis}\label{sec:ablation}

\noindent We include here additional analyses and experiments which are useful to motivate methodological choices, while also presenting an extended ablation for the relevant hyper-paramters. More in detail we analyze:

\begin{itemize}
\item The bias identification approaches: we compare the subdivision by Prediction History (PH) with the split obtained by Single Prediction (SP).
\item How the quality of the subdivision of the data in biased/unbiased samples performed by the initial pseudo-labeling stage affects the final accuracy.
\item How learning the Beta distribution parameters impacts the performance, namely, we compare our proposed \textit{l-mix} with the naive \textit{s-mix}.
\item \textit{s-mix} performances by varying the parameter $\zeta$ controlling the contribution of the regularization term.
\item \textit{l-mix} performances by varying the parameter $\omega$ controlling the contribution of the regularization term.
\item The possible augmentation policies (sampling strategies) and how these affects the performance at the different bias ratios.
\item An application of the proposed methods to an \textbf{unbiased} dataset showing that they are still beneficial in this scenario, i.e., generalization is improved with respect to standard ERM even with no apparent bias.
\end{itemize}

To have the full control of the experimental conditions, we conducted most of these trials and the ablation analysis using Corrupted CIFAR-10 (bias ratio$=95\%$) if not differently specified, in order to assess the contribution of each step characterizing our approach.

\subsection{Quality of Bias Identification}
We assessed the quality of the split obtained by the Prediction History (PH) method. We consider F1-score, Precision and Recall as metrics to evaluate the discrepancy between the pseudo-labels and the ground-truth bias/unbias split (see Fig. \ref{fig:ablation_gamma}).  
CIFAR-10 data (bias ratio $95\%$) is considered. We evaluated the method over different values of $M$, namely the amount of epochs before refreshing the assignment: we can notice that as long as $M$ increases, recall decreases and precision increases, while F1-score is significantly higher than the random split case (see Table \ref{table_f1_score}). We chose $M=5$ for all our experiments, while different values are not affecting significantly the performance.

\subsection{Effect of Bias Identification on model performance}

We investigate how different amounts of noise in the pseudo-labeling stage impact the final outcome in terms of test accuracy. We employed our \textit{l-mix} method on CIFAR-10, see Table \ref{table_f1_score}. We show the test accuracies in the ideal case of perfect subdivision between biased and unbiased samples (Oracle, $F1=1$), by applying our approach on top of Single Prediction and Prediction History procedures, and in the case of random split. We noted that passing from the oracle conditions (best) to the random split (worst), accuracy drops significantly. Interestingly, the method trained on the random pseudo-labeling always achieves better accuracy than that of the ERM baseline (see Table \ref{table_toy_test}), and even higher than those achieved by some methods specifically designed to debias data. Even a coarse split (better than random) considerably increases the final performance with respect to ERM training. Also, our mixing strategy has a strong regularization effect even in suboptimal conditions (i.e., when mixing mostly biased samples). Comparing with the two bias identification strategies SP and PH, we see that the difference is not severe (drop between 1-4\% for Single Prediction).


\begin{table*}[h!]
\small
\begin{center}
{
\begin{tabular}{l|cc|cc|cc|cc}
\toprule

& \multicolumn{2}{c|}{\textbf{Oracle}} & \multicolumn{2}{c|}{\textbf{Random split}} & \multicolumn{2}{c|}{\textbf{SP}} & \multicolumn{2}{c}{\textbf{PH}} \\

\midrule

Bias ratio & F1 & Test Acc. & F1 & Test Acc. & F1 & Test Acc. & F1 & Test Acc.\\ 

\midrule

95\%   & 1.0 & 66.3 & 0.37 & 50.3 & 0.65 & 60.9 & 0.74 & \textbf{64.6}\\

98\%   & 1.0 & 59.4 & 0.34 & 40.4 & 0.62 & 54.4 & 0.67 & \textbf{57.2}\\

99\%   & 1.0 & 54.7 & 0.33 & 29.7 & 0.58 & 49.3 & 0.63 & \textbf{50.6}\\

99.5\% & 1.0 & 49.0 & 0.32 & 21.5 & 0.54 & 42.5 & 0.59 & \textbf{43.7}\\

\bottomrule

\end{tabular}}
\end{center}
\caption{ \textbf{Ablation study on bias identification on Corrupted CIFAR-10.} The table reports F1 scores for $\hat{\mathcal{D}}_{bias}$ / $\hat{\mathcal{D}}_{unbias}$ splitting, obtained with our bias identification strategies Single Prediction (SP) and Prediction History (PH) presented in Section \ref{sec:BI}. We compare them to oracle (groundtruth) and randomly generated subsets. We report the final test accuracy of \textit{l-mix} on the whole test set for the four different cases. Best accuracies are obtained with oracle splitting, while worst for random splitting. PH performs very favourably with respect to the oracle in terms of test accuracy, even if the split is not perfect.
}
\label{table_f1_score}
\end{table*}
\begin{table}[h!]
\centering

\begin{tabular}{l|cc|cc|cc}
\toprule
$\omega$ &  \multicolumn{2}{c}{\textbf{Corrupted CIFAR-10}} & \multicolumn{2}{c}{\textbf{Waterbirds}} & \multicolumn{2}{c}{\textbf{CelebA}}\\
& All & Unbiased & All & Unbiased & All & Unbiased \\
\midrule
$0$       & 61.8 & 62.0 &  91.6 & 84.3 & 85.4 & 78.9 \\ 
$10^{-4}$ & 63.7 & 63.8 &  92.6 & 86.2 & 87.1 & 80.6 \\ 
$10^{-3}$  & \textbf{64.7} & \textbf{64.6} &  \textbf{94.7} & \textbf{87.7} & \textbf{89.1} & \textbf{82.2} \\
$10^{-2}$ & 61.9 & 62.2 &  91.9 & 85.1 & 86.7 & 80.1 \\ 
$10^{-1}$ & 60.6 & 61.5 &  89.6 & 82.3 & 83.8 & 75.7 \\ 
\bottomrule
\end{tabular}

\caption{Test accuracy for Unbiased samples and All (full test set) for different values of $\omega$ on \textit{l-mix}, evaluated on different benchmarks. The values for $\omega$ are chosen on a logarithmic scale. 
}
\label{tab_regularizer}
\end{table}
\begin{table*}
\centering
\begin{tabular}{ll|cccc|c}

\toprule

 \multirow{2}{*}{\textbf{Set 1}}  & \multirow{2}{*}{\textbf{Set 2}}  & \multicolumn{4}{c|}{\textbf{Bias ratio}} & \\

        &        & 95\% & 98\% & 99\% & 99.5\%\\

\midrule

\multicolumn{2}{c|}{No augmentation} & 58.8\% & 46.1\%  & 40.0\%  & 33.6\% &\multirow{5}{*}{\rotatebox[origin=c]{90}{All}} \\

\cmidrule(r){1-2}

\multicolumn{2}{c|}{Vanilla MixUp} & 44.2\% & 39.4\% & 37.2\% & 35.2\% &\\

\cmidrule(r){1-2}

$\hat{\mathcal{D}}_{bias}$ & $\hat{\mathcal{D}}_{bias}$     &  35.2\%    & 34.0\%  & 32.9\% & 32.0\% & \\

$\hat{\mathcal{D}}_{unbias}$ & $\hat{\mathcal{D}}_{unbias}$ &  60.2\%  & 54.1\% & 48.4\% & 40.4\%  & \\


$\hat{\mathcal{D}}_{bias}$ & $\hat{\mathcal{D}}_{unbias}$   &  \textbf{63.8\%}  & \textbf{56.4\%} & \textbf{50.9\%}   & \textbf{43.1\%} & \\

\bottomrule
\end{tabular}

\begin{tabular}{ll|cccc|c}

\multicolumn{2}{c|}{No augmentation}  & 55.3\% & 41.5\%  & 34.8\%  & 27.1\% &\multirow{5}{*}{\rotatebox[origin=c]{90}{Unbiased}} \\

\cmidrule(r){1-2}

\multicolumn{2}{c|}{Vanilla MixUp} & 42.1\% & 36.7\% & 36.5\%  & 34.4\% &\\

\cmidrule(r){1-2}

$\hat{\mathcal{D}}_{bias}$ & $\hat{\mathcal{D}}_{bias}$     &  29.7\%  & 28.4\% & 26.7\% & 27.5\% &\\


$\hat{\mathcal{D}}_{unbias}$ & $\hat{\mathcal{D}}_{unbias}$ &  63.1\% & 55.3\% & 48.7\%  & 42.5\% &\\


$\hat{\mathcal{D}}_{bias}$ & $\hat{\mathcal{D}}_{unbias}$    &  \textbf{63.3\%} & \textbf{55.9\%} & \textbf{49.4\%}  & \textbf{42.7\%} & \\

\bottomrule
\end{tabular}

\caption{\textbf{Ablation analysis on the augmentation strategies.} For Corrupted CIFAR-10, we report the accuracy resulting from different augmentation strategies and no augmentation, by varying the bias ratio. Our strategy results the winner over all the other mixing policies for both the entire test set (\textit{All}) and the \textit{Unbiased} portion (best results are in bold). 
}
\label{table_ablation_data_v2}
\end{table*}

\subsection{\textit{l-mix} vs. \textit{s-mix}}
In Table \ref{table_ablation_s-l}, we analyze how \textit{learning} the parameters of the $Beta$ distribution results in improved performance on both the whole test test and the unbiased subset. The improvement is consistent across all datasets analyzed. We set $\zeta=10$ for the standard Mixup \cite{mixup} strategy we named \textit{s-mix}, and $\omega=10^{-3}$ for \textit{l-mix} (check the detailed ablation study fr the two parameters in \ref{subsec-zetaablation} and \ref{subsec-omegaablation} respectively). 
\begin{table*}[h!]
\small
\begin{center}
{
\begin{tabular}{l|cc|cc|cc|c}
\toprule

& \multicolumn{2}{c|}{\textbf{CIFAR-10 (95\%)}} & \multicolumn{2}{c|}{\textbf{Waterbirds}} & \multicolumn{2}{c|}{\textbf{CelebA}} & \textbf{BAR} \\

\midrule

& All & Unbiased & All & Unbiased & All & Unbiased & All  \\ 

\midrule

\textit{s-mix}  & 63.3 & 63.3 & 94.3 & 87.1 & 88.3   & 81.6 & 64.3 \\

\textit{l-mix}  & \textbf{64.7} & \textbf{64.6} & \textbf{94.7} & \textbf{87.7}  & \textbf{89.1} & \textbf{82.1} & \textbf{64.8} \\

\bottomrule

\end{tabular}
}
\end{center}
\caption{Ablation study on learnable vs simple mixing strategy. For \textit{l-mix} the parameters $\alpha$ and $\beta$ of the Beta distribution are learned by the neural network $h$ (eq. \ref{eq:alphabeta}) via equation \ref{eq:optimi}; for \textit{s-mix} they are kept fixed as in \cite{mixup}. We set $\zeta=10$ for \textit{s-mix and} $\omega=10^{-3}$ for \textit{l-mix}.
}
\label{table_ablation_s-l}
\end{table*}

\begin{figure}[!t]
    \centering
    \includegraphics[width=0.75\linewidth]{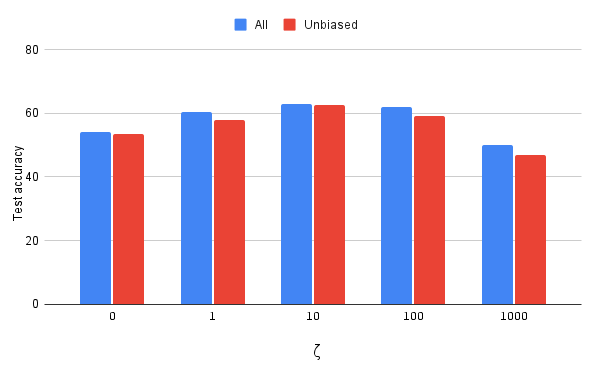} 

    \caption{Test accuracy for unbiased samples (red) and full set of samples (blue) for different values of $\zeta$. For $\zeta=0.0$, the loss corresponds to the weighted ERM of Eq. \ref{eq:loss_method1}. Note that The $\zeta$ axis is in logarithmic scale. 
    }
    \label{fig:ablation_zeta}
\end{figure}

\label{subsec-zetaablation}
\noindent We conducted an analysis on the $\zeta$ hyperparameter of \textit{s-mix}  (Eq. \ref{eq:loss_method1}). We investigated $\zeta$ in the range $[0, 10^{-4}, 10^{-3}, 10^{-2}, 10^{-1}]$ and evaluated the performance on Corrupted CIFAR-10 on both biased and unbiased test samples, see Fig. \ref{fig:ablation_zeta}. Even in this case, we observe a large range in which the method is benefiting from the regularization effect of Eq. \ref{eq:loss_method1}. The case $\zeta=0$ represents the final loss in which the regularizer contribution is canceled out. 
As reported in Sect. \ref{sec:experiments}, we always used $\zeta=10$ for evaluating \textit{s-mix} performances.

\subsection{Ablation study on $\omega$ for \textit{l-mix}}
\label{subsec-omegaablation}
\noindent Similarly, we set different values of the parameter $\omega$ of Eq. \ref{eq:loss_method2}, i.e., $[0, 10^{-4}, 10^{-3}, 10^{-2}, 10^{-1}]$, and evaluated the performance of \textit{l-mix} on Corrupted CIFAR-10 (bias ratio $=95\%$). In Table \ref{tab_regularizer}, we report the accuracy on the unbiased and on the full set of samples.
We can note that there is a large range of $\omega$ values,  $10^{-4} < \omega < 10^{-2}$, in which the accuracy is reaching high values in both cases (please, note the logarithmic scale in the $x$ axis). This empirically shows that $\omega$ is not so a sensitive parameter with respect to the proposed strategy, and for this reason, we fix $\omega=10^{-3}$ in all our experiments.

\subsection{Augmentation strategies}

We report in Table \ref{table_ablation_data_v2} an ablation study on the augmentation strategies, consisting in sampling $\hat{x}_1$, $\hat{x}_2$ in different ways from either $\hat{\mathcal{D}}_{bias}$ or $\hat{\mathcal{D}}_{unbias}$ or both. The study was carried on as a preliminary investigation on \textit{s-mix} only, in order to decouple the sample strategy from the learning problem.
We consider the Corrupted CIFAR-10 benchmark, and show the final accuracies for the unbiased and full sets, for all the bias ratios considered. Sampling both $\hat{x}_1$, $\hat{x}_2$ from $\hat{\mathcal{D}}_{bias}$ overfits the biased data and results in the worst accuracy, while mixing both samples from $\hat{\mathcal{D}}_{unbias}$ increases the generalization over unbiased samples, but provides suboptimal results, especially for the biased test set. Mixing samples from $\hat{\mathcal{D}}_{bias}$ and $\hat{\mathcal{D}}_{unbias}$ corresponds to our strategy, which provides the best performance. 

We also report the baseline case in which no augmentation is performed (1st and 6th rows), i.e. $x_{mix}$, $y_{mix}$ in equations \ref{eq:mixup} and  \ref{eq:loss_method1} are just individual samples randomly drawn from $\mathcal{D}$: performances in this case are significantly distant from those obtained by our proposed approaches. 
We also considered the vanilla Mixup data-augmentation as a baseline (2nd and 7th rows). Results are slightly higher than those of mixing $\hat{\mathcal{D}}_{bias}$ -  $\hat{\mathcal{D}}_{bias}$ (3rd and 8th rows), but lower than the other combinations. This was somehow expected as most of the samples in the dataset are biased, therefore it is very likely that, by selecting pairs to be mixed randomly, biased/biased pairs are picked more often, while biased/unbiased or unbiased/unbiased pairs are less frequently  considered. 

To conclude, this analysis proves that by choosing pairs from $\hat{\mathcal{D}}_{bias}$ and $\hat{\mathcal{D}}_{unbias}$ is much more effective, even when the two splits are not accurate.

\subsection{Effect on unbiased datasets}

\begin{wraptable}{r}{4cm}

\begin{tabular}{lcc}

\toprule

\multicolumn{3}{c}{\textbf{Test accuracy}} \\

\textbf{ERM} & \textbf{\textit{s-mix}} & \textbf{\textit{l-mix}} 
\\

\midrule

92.94 & 93.65 & \textbf{94.02}\\

\bottomrule

\end{tabular}
\caption{ Ablation study on non-biased CIFAR-10. }
\label{tab:tab_non_biased}
\end{wraptable}

To conclude, we tested our proposed methods on the standard CIFAR-10 dataset, without any induced bias. With this test, we want to prove that even in the presence of a non-biased dataset, our proposed method is beneficial, showing an improved generalization capacity. In other words, we can tackle a real-world unsupervised debiasing problem, in which we do not actually know a priori whether a dataset is biased or not. We expect that our proposed methods do not underperform with respect to standard ERM training.

Differently from Corrupted CIFAR-10, the standard version of the dataset does not exhibit any known shortcut when it comes to predict the class labels. We compared our proposed method \textit{l-mix} and the non-learned version \textit{s-mix} with standard ERM, using the same architecture $f_{\theta}$ as backbone for a fair comparison (ResNet-18). The dataset split is performed with the PH approach ($M=5$). Results are reported in Table \ref{tab:tab_non_biased}, which reports the test accuracy of a plain ResNet-18 trained via ERM on CIFAR-10, compared with \textit{s-mix} (regularization parameter  $\zeta=10$, and \textit{l-mix}, with regularization parameter $\omega=10^{-3}$).
We can observe that in both cases, our results exceed the ERM baseline, thus showing that our approaches are agnostic to the fact that the dataset might be biased or not, making them more amenable to be utilized in real use cases. 

\section{Conclusions}
\label{sec:conclusion}
In this work, we address the problem of bias in the data in the realistic, unsupervised scenario, i.e., when the bias factor or attribute is unknown, and propose a novel approach to mitigate this issue. Our approach is composed of two sequential stages. First, we introduce two strategies to subdivide the training dataset into two subsets (biased and unbiased samples), namely SP and PH. Relying on the history of predictions, PH results in cleaner separation of biased/unbiased samples, which in turn results in better performance for the subsequent bias removal stage.

Second, we show how to effectively exploit such subdivision in order to produce augmented samples by mixing estimated biased andunbiased samples by properly learning mixing coefficients in an adversarial way. Such mixed samples act as a regularizer which breaks the spurious correlations between data and class labels, thus increasing accuracy. 

The overall approach scores favorably against state of the art approaches and is also quite robust to the choice of hyper-parameters, for which we present a comprehensive ablation analysis. Notably, the method is completely agnostic of the presence of bias, and it also outperforms standard ERM even when no bias is (apparently) present in the data.

\backmatter

\bibliography{sn-article-template/biblio}

\end{document}